\documentclass[review]{elsarticle}

\usepackage{amsmath} 
\usepackage{graphicx} 
\usepackage[table,xcdraw]{xcolor} 
\usepackage[utf8]{inputenc} 	
\usepackage{multirow}
\usepackage[normalem]{ulem}

\usepackage{tikz}
\usepackage{textcomp}
\usepackage{lipsum}

\journal{Neural Networks}

\begin{document}
	
	\begin{frontmatter}			
		
		\title{On the Compression of Neural Networks Using $\ell_0$-Norm Regularization and Weight Pruning}
		
		\author[linseaddress]{Felipe~Dennis~de~Resende~Oliveira}
		\author[linseaddress]{Eduardo~Luiz~Ortiz~Batista\corref{mycorrespondingauthor}}
		\ead{eduardo.batista@ufsc.br}
		\author[linseaddress]{Rui~Seara}

		\address[linseaddress]{LINSE—Circuits and Signal Processing Laboratory, Department of Electrical Engineering, Federal University of Santa Catarina, Florianópolis, 88040-900, Brazil}
		
%
		\cortext[mycorrespondingauthor]{Corresponding author}
		
%

		\begin{abstract}						
			Despite the growing availability of high-capacity computational platforms, implementation complexity still has been a great concern for the real-world deployment of neural networks. This concern is not exclusively due to the huge costs of state-of-the-art network architectures, but also due to the recent push towards edge intelligence and the use of neural networks in embedded applications. In this context, network compression techniques have been gaining interest due to their ability for reducing deployment costs while keeping inference accuracy at satisfactory levels. The present paper is dedicated to the development of a novel compression scheme for neural networks. To this end, a new form of $\ell_0$-norm-based regularization is firstly developed, which is capable of inducing strong sparseness in the network during training. Then, targeting the smaller weights of the trained network with pruning techniques, smaller yet highly effective networks can be obtained. The proposed compression scheme also involves the use of $\ell_2$-norm regularization to avoid overfitting as well as fine tuning to improve the performance of the pruned network. Experimental results are presented aiming to show the effectiveness of the proposed scheme as well as to make comparisons with competing approaches.
		\end{abstract}
		

	\end{frontmatter}

	

\section{Introduction}
The importance of neural networks in the machine learning and engineering contexts has been growing  substantially over the last decades \cite{goodfellow2016deep,bishop2006pattern, murphy2013machine}. 
Factors contributing to this growth include notable improvements in the training algorithms and the increasing availability of high-capacity computational platforms, which have allowed the deployment of neural networks having larger numbers of internal layers. Such networks, termed deep neural networks (DNNs), have been successfully applied to a number of complex problems, such as the stock market prediction \cite{akita2016deep, vargas2017deep, tsantekidis2017forecasting}, sentiment analysis \cite{jianqiang2018deep, hassan2017deep},  natural language processing \cite{young2018recent, klosowski2018deep}, speech recognition \cite{nassif2019speech, liao2015towards, kosaka2015deep}, and  image processing \cite{lecun1998gradient, bandhu2017classifying, krizhevsky2012imagenet}.

Despite the aforementioned increasing computational power availability, computational costs still pose an important challenge for the use of DNNs in real-world applications. 
This point is due to the fact that many effective DNNs deal with a huge number of parameters (weights).  
One particular example is AlexNet (winner of the 2012 Imagenet Large Scale Visual Recognition Challenge) \cite{krizhevsky2012imagenet}, which comprises more than 60 million learnable weights. The training of networks with so many parameters, in spite of being computationally costly and time consuming, is a more tractable problem since it is typically an offline process. In contrast, deployment costs tend to be more critical and often prohibitive for large networks, especially in applications involving real-time processing and/or embedded systems.
Aiming to address this hindrance, a wide range of research works have been dedicated to the development of network compression techniques. Some of these techniques are based on using matrix decompositions to find suitable network approximations with reduced computational complexity \cite{NIPS2013_5025, Sainath2013, Zhang2015, Sun2017}. Other techniques exploit hardware implementation characteristics to achieve computational cost reductions \cite{Vincent2011, Hwang2014, Anwar2015}. Moreover, neuron- or weight-pruning approaches have also been considered for developing network compression techniques \cite{NIPS1989_250, NIPS1992_647, DBLP:journals/corr/GuoYC16, Yu2012, Mauch2017, srinivas2015data}. Beyond this list of examples, the reader can refer to one of the recent survey papers to obtain a more comprehensive perspective on the field of neural network compression \cite{survey1liang,methodsVadera}.

Specifically in the context of network compression via pruning approaches, strategies that combine norm-penalization-based regularization with pruning have been gaining significant attention \cite{NIPS2015_5784, pan2016dropneuron, louizos2017learning, scardapane2017group}. This aspect is due to the fact that such regularization techniques tend to induce network sparseness, which potentially increases the effectiveness of a subsequent weight and/or neuron pruning. 
In \cite{NIPS2015_5784}, for instance, the association of $\ell_2$-norm regularization with weight pruning has achieved up to 12 times reductions on the number of network weights without affecting its performance.
Moreover, in \cite{scardapane2017group}, 38 times reductions have been obtained at the cost of small performance losses using a combination of $\ell_1$-norm regularization and weight pruning. When neuron pruning is desired, group-based regularization techniques have led to promising results \cite{scardapane2017group, louizos2017learning}. Moreover, as shown in \cite{pan2016dropneuron}, it is possible to combine neuron and weight penalization to increase the network compression. 

The focus of the present research work is on the development of a new strategy for neural network compression based on norm regularization and weight pruning. The idea here is to use the $\ell_0$ norm instead of the commonly used $\ell_1$ or $\ell_2$ norms, since $\ell_0$-norm penalization typically has a stronger sparseness-inducing effect, as discussed in \cite{l0sparseinducting}. 
Other types of $\ell_0$-norm regularizations have also been applied to network compression in \cite{louizos2017learning, li2019l0, zhangECCV2018, cperpinanL0, cperpinanL0L2}. Each of these works adopts a different approach  to deal with the difficulties arising from the use of the non-differentiable $\ell_0$-norm function. In \cite{louizos2017learning}, for instance, a general framework for using surrogate $\ell_0$ functions for regularization is introduced together with an $\ell_0$-norm approximation, which is termed hard-concrete.
The strategy introduced in \cite{li2019l0} is built on top of the framework given in \cite{louizos2017learning}, replacing the hard-concrete-based gradient estimator by the so-called augment-reinforce-merge (ARM) estimator, which has allowed obtaining improved results. The focus in both \cite{louizos2017learning} and \cite{li2019l0} is mostly on exploiting group sparseness aiming at carrying out neuron pruning. In \cite{zhangECCV2018}, the weight pruning is formulated as a constrained nonconvex optimization problem having a constraint on the cardinality of the weights (i.e., on the $\ell_0$ norm of the weight vector). The alternating direction method of multipliers (ADMM) is used for systematically solving such a problem and the weight pruning is then carried out by removing the weights that are close to zero \cite{zhangECCV2018}. In \cite{cperpinanL0}, the weight pruning problem is formulated in two different forms: as a constrained optimization  (with a $\ell_0$-norm constraint, similar to \cite{zhangECCV2018}) and also as an optimization problem with a penalized objective function. Alternating optimization approaches are used to solve the problems obtained in the two proposed forms, leading to two-step algorithms for network training with weight pruning \cite{cperpinanL0}. Moreover, in \cite{cperpinanL0L2}, the approach introduced in \cite{cperpinanL0} is used for combining the $\ell_0$ and $\ell_2$ norms as constraints/penalizations, leading to improved results.

Contributing to this research topic, a new weight pruning approach that uses $\ell_0$-norm regularization is introduced in the present work. The proposed approach is based on defining an optimization problem with an $\ell_0$-norm-based penalization on the objective function, bearing some similarity the traditional $\ell_2$ and $\ell_1$ norm regularizations used for overfitting mitigation and also with the approach from \cite{cperpinanL0}. To make such a problem more tractable, we rely on the exponential approximation of the $\ell_0$ norm discussed in \cite{gu2009l_}. In doing so, a gradient-descent-based strategy can be used for obtaining an effective network training algorithm, avoiding the use of more complex optimization approaches such as the ones based on alternating optimization. We also analyze the zero-attracting effect provided by the obtained training algorithm, showing that, for small weights, this effect is stronger than that obtained using $\ell_2$- and $\ell_1$-norm regularizations. However, for larger weights, such a zero-attracting effect tends to zero, making the proposed training algorithm not effective for avoiding model overfitting. To circumvent this problem, we propose a combined $\ell_2$-$\ell_0$-norm regularization in which the $\ell_2$-norm penalization is responsible to avoid overfitting, whereas the $\ell_0$-norm one targets smaller weights aiming at sparseness induction. This approach differs from the one in \cite{cperpinanL0L2}, where the $\ell_2$-norm regularization aims at increasing the zero-attracting effect. The proposed combined $\ell_2$-$\ell_0$-norm regularization in association with weight pruning gives rise to the network compression scheme proposed in this paper. A set of experimental results are presented aiming to show the effectiveness of the proposed compression scheme as well as to make comparisons with other competing strategies.

The remainder of this paper is organized as follows. Section \ref{sec:regularizations} presents a review of classic regularization techniques based on vector norms. The proposed approach is discussed in Section \ref{sec:desenv}. Finally, Sections \ref{sec:results} and \ref{sec:conc} are dedicated to present, respectively, experimental results and concluding remarks.

	%
	
	
\section{Background on Standard Norm-Based Regularization and Its Application to Weight Pruning}
\label{sec:regularizations}

In general terms, norm-based regularization strategies aim at preventing the overfitting by means of introducing a norm-dependent penalization into the training process \cite{goodfellow2016deep}. This operation is carried out by considering an objective function in the following form to derive the network training algorithm:
\begin{equation}
\tilde{J}(\mathbf{w}, \mathbf{x}, \mathbf{y}) = J(\mathbf{w}, \mathbf{x}, \mathbf{y}) + \alpha \Omega(\mathbf{w})
\label{eq:cost_func_penalities}
\end{equation}
where $J(\mathbf{w}, \mathbf{x}, \mathbf{y})$ corresponds to a standard (non-regularized) objective function, $\Omega(\mathbf{w})$ is the penalization function, and $\alpha$ is a parameter that allows controlling the penalization level. Note that $\Omega(\mathbf{w})$ is a function exclusively of the weight vector $\mathbf{w}$ (containing all network weights), whereas $J(\mathbf{w}, \mathbf{x}, \mathbf{y})$ depends also on $\mathbf{x}$ and $\mathbf{y}$, which correspond, respectively, to the input and output data from the training dataset.

The minimization of \eqref{eq:cost_func_penalities} is generally pursued by means of adjusting $\mathbf{w}$ using gradient-descent-based methods \cite{goodfellow2016deep}. Thus, the update expression for the $j$th weight ($w_j$) can be represented as
\begin{equation}\label{eq:update_weights}
w_j^{+} \leftarrow w_j - \eta \frac{\partial \left[J(\mathbf{w}, \mathbf{x}, \mathbf{y}) + \alpha \Omega(\mathbf{w}) \right]}{\partial w_j}
\end{equation}
with $\eta$ denoting the learning rate and $w_j^{+}$, the updated (\textit{a posteriori}) value of $w_j$. 
Note that, for $\alpha = 0$, \eqref{eq:update_weights} becomes the standard (non-regularized) weight-update rule, which is given by
\begin{equation}\label{eq:updatestd}
w_j^{+} \leftarrow w_j - \eta \frac{\partial J(\mathbf{w}, \mathbf{x}, \mathbf{y})}{\partial w_j}.
\end{equation}

\subsection{$\ell_2$-Norm Regularization}
One of the most popular regularization techniques that use norm penalties is the $\ell_2$-norm regularization. This technique is developed by making  
\begin{equation}\label{eq:norm_l2}
\Omega(\mathbf{w}) = ||\mathbf{w}||^2_2 = \sum_{j \in \phi} w_j^2
\end{equation}
in \eqref{eq:cost_func_penalities} \cite{goodfellow2016deep},
with $\phi$ representing the set of penalized network parameters. 
By substituting \eqref{eq:norm_l2} into \eqref{eq:update_weights}, considering that $\partial ||\mathbf{w}||^2_2 / \partial w_j = 2 w_j $, and manipulating the resulting expression, 
the following update equation is obtained:
\begin{equation}\label{eq:update_j_weight_2}
w_j^{+} \leftarrow w_j - 2\eta\alpha w_j - \eta \frac{\partial J(\mathbf{w}, \mathbf{x}, \mathbf{y})}{\partial w_j}.
\end{equation}

By comparing \eqref{eq:update_j_weight_2} with the standard (non-regularized) update equation given in \eqref{eq:updatestd}, one can notice that the $\ell_2$-norm regularization introduces an extra term, given by $-2\eta\alpha w_j$, into the weight update equation. Since $2\eta\alpha$ is always positive and generally much smaller than one (for the sake of algorithm convergence), this extra term produces a reduction of $2\eta\alpha |w_j|$ in the weight magnitude, which corresponds to an attraction of the weight towards zero. Due to this characteristic, such a term will be referred to as zero-attraction term in the present research work. The weight penalization produced by the zero-attraction term is commonly used to avoid the overfitting \cite{goodfellow2016deep}.

\subsection{$\ell_1$-Norm Regularization}
In the case of the $\ell_1$-norm regularization, the penalization function is given by
\begin{equation}\label{eq:norm_l1}
\Omega(\mathbf{w}) = ||\mathbf{w}||_1 = \sum_{i \in \phi} |w|_i.
\end{equation} 
Then, considering \eqref{eq:update_weights}, \eqref{eq:norm_l1}, and also that $\partial ||\mathbf{w}||_1 / \partial w_j = \mathrm{sign}(w_j)$, the following weight-update equation is obtained:
\begin{equation}\label{eq:update_weights_l1}
w_j^{+} \leftarrow w_j - \eta\alpha \, \mathrm{sign}(w_j) - \eta \frac{\partial J(\mathbf{w}, \mathbf{x}, \mathbf{y})}{\partial w_j}.
\end{equation}
Note that \eqref{eq:update_weights_l1} corresponds to \eqref{eq:updatestd} with the inclusion of the term $-\eta\alpha \, \mathrm{sign}(w_j)$, which is also a zero-attraction term since $0 < \eta\alpha \ll 1$.

\subsection{Network Compression via Norm-Based Regularization and Weight Pruning}\label{sec:associating}

As previously mentioned, both  $\ell_2$- and $\ell_1$-norm regularizations result in weight-update zero-attraction terms that induce weight magnitude reductions during the network training. Such a zero-attraction effect, besides preventing model overfitting, also tends to produce networks with several weights having very small magnitude values. Most of these weights have a negligible contribution to the network output and, thus, they can be pruned out. This idea is the basis for many network compression techniques, in which a magnitude threshold is considered for pruning weights after a norm-regularized training process \cite{NIPS2015_5784, louizos2017learning, li2019l0, xie2019l0}.

	%
	
	\section{Proposed Scheme}\label{sec:desenv}	

In this section, the focus is on the main contributions of the present paper. The first is an $\ell_0$-norm-based regularization approach for network training, which aims at producing networks having larger amounts of prunable weights. Then, a discussion is presented regarding the zero-attraction characteristics of the proposed $\ell_0$-norm-based approach as compared with the other norm-based regularization approaches. From this discussion, a scheme combining both $\ell_2$- and $\ell_0$-norm regularizations emerges, which is associated with weight pruning to give rise to the proposed network compression scheme.

\subsection{Proposed $\ell_0$-Norm-Based Regularization Approach\label{sec:proposedl0}}
The $\ell_0$ norm is well known for being a strict measure of vector sparseness, since it is formally defined as the number of nonzero coefficients in a vector \cite{louizos2017learning, gu2009l_, mancera2006l0}. Due to this fact, the $\ell_0$ norm has been considered in applications involving optimization and sparseness induction \cite{louizos2017learning, li2019l0, gu2009l_, xie2019l0, mancera2006l0}, which share important similarities with the network compression strategies of interest in the present research work. Thus, aiming to develop a novel regularization approach, let us start with the formal definition of the $\ell_0$ norm of a weight vector $\mathbf{w}$, which is given by
\begin{equation}\label{eq:norma_l0}
	||\mathbf{w}||_0 = \sum_{j \in \phi} f(w_j)
\end{equation} 
with
\begin{equation}\label{eq:func_norma_l0}
	f(w_j) = \left \{ 
	\begin{matrix}
	1 \text{ if } w_j \neq 0\\
	0 \text{ otherwise.}\\
	\end{matrix}
	\right.
\end{equation}
Note that \eqref{eq:norma_l0} is a non-convex and non-differentiable function. These characteristics pose important difficulties for the development of regularization strategies as those described in Section~\ref{sec:regularizations}, especially due to the central role of gradient descent optimization methods [see \eqref{eq:update_weights}]. 
To circumvent this problem, we consider here an approximate differentiable version of the $\ell_0$ norm described in \cite{gu2009l_}. More specifically, \eqref{eq:func_norma_l0} is approximated by
\begin{equation}\label{eq:aprox_l0}
f^*(w_j) = 1 - e^{-\beta|w_j|},
\end{equation}
with $\beta \geq 1$. Thus, using \eqref{eq:aprox_l0} in lieu of $f(w_j)$ in \eqref{eq:norma_l0}, the following $\ell_0$ norm approximation is obtained:
\begin{equation}\label{eq:norma_l0_aprox}
||\mathbf{w}||_0 \cong ||\mathbf{w}||_{0}^* =  \sum_{j \in \phi} (1-e^{-\beta|w_j|}).
\end{equation}
Parameter $\beta$ determines the degree of similarity between the approximation and the $\ell_0$ norm (i.e., as $\beta \to \infty$, \mbox{$ ||\mathbf{w}||_0^* \to ||\mathbf{w}||_0$}). 

By considering $\Omega(\mathbf{w}) = ||\mathbf{w}||_{0}^*$ in \eqref{eq:update_weights}, we have
\begin{equation}\label{eq:update_weight_aprox_l0}
	w_j^{+} \leftarrow w_j - \eta \alpha \frac{\partial || \mathbf{w}||_0^*}{\partial w_j} - \eta \frac{\partial J(\mathbf{w}, \mathbf{X}, \mathbf{y})}{\partial w_j}.
\end{equation}
Now, taking the gradient of \eqref{eq:norma_l0_aprox} with respect to $w_j$, we get
\begin{equation}\label{eq:l0normgradient}
	\frac{\partial ||\mathbf{w}||_0^*}{\partial w_j} = \beta \, \mathrm{sign}(w_j) \, e^{-\beta |w_j|}.
\end{equation}
Finally, substituting \eqref{eq:l0normgradient} into \eqref{eq:update_weight_aprox_l0}, the following $\ell_0$-norm-regularized weight update rule is obtained:
\begin{equation}\label{eq:update_weight_aprox_l0_2}
	w_j^{+} \leftarrow w_j - \eta \alpha \beta \mathrm{sign}(w_j) e^{-\beta|w_j|} - \eta \frac{\partial J(\mathbf{w}; \mathbf{X}, \mathbf{y})}{\partial w_j}.
\end{equation}
Note that \eqref{eq:update_weight_aprox_l0_2} corresponds to \eqref{eq:updatestd} with an additional term given by \linebreak $- \eta \alpha \beta \mathrm{sign}(w_j) e^{-\beta|w_j|}$, which also corresponds to a zero-attraction term since in practice \mbox{$ 0 \leq \eta \alpha \beta e^{-\beta|w_j|} \ll 1$}.

\subsection{Zero-Attraction Comparison for Different Norm-Based Regularization Strategies}

\begin{figure}[tb]
	\centering
	\includegraphics[scale=0.9]{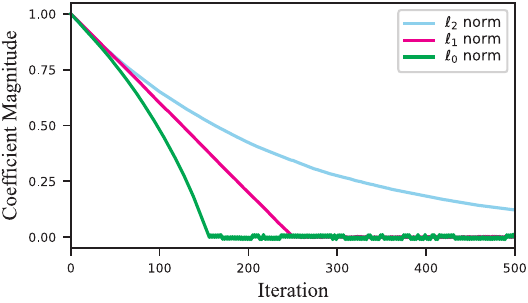}
	\vspace{-0.5cm}
	\caption{Behavior obtained for a weight initially set to $1.0$ and subjected to the different norm regularizations with similar initial zero-attraction rate.}
	\label{fig:zeroattractillustration}
\end{figure}

As mentioned in Section~\ref{sec:associating}, the zero-attraction terms present in \eqref{eq:update_j_weight_2} ($\ell_2$-norm-regularized weight update) and \eqref{eq:update_weights_l1} ($\ell_1$-norm-regularized weight update) exert a magnitude reduction over the network weights during training. This point is also the case for the zero-attraction term in the $\ell_0$-norm-regularized weight update  given in \eqref{eq:update_weight_aprox_l0_2}. The characteristics of such zero attractions are however significantly different between each other. By comparing them (all listed together in Table~\ref{tab:zeroatchars}), one can notice that: i) the $\ell_2$-norm zero attraction depends on $w_j$ and thus on the weight magnitude \cite{goodfellow2016deep}, which results in a decreasing attraction as the weight approaches zero; ii) the $\ell_1$-norm zero attraction depends on the weight sign (not on its magnitude) resulting in a steady attraction effect \cite{goodfellow2016deep}; and iii) the $\ell_0$-norm-based zero attraction depends on the weight sign as well as on its magnitude via $e^{-\beta|w_j|}$, resulting in an attraction that in fact increases as the weight approaches zero. These characteristics are evidenced by the curves shown in Figure~\ref{fig:zeroattractillustration}, which depict the behavior of a weight subject to the distinct norm regularizations. These curves have been obtained setting the initial weight value to $1.0$ and adjusting the parameters of the different regularizations in order to obtain approximately the same initial zero-attraction rate.

\begin{table}[b]
	\vspace{-.5cm}
	\caption{Zero-attraction characteristics for the different norm-based regularizations}
	\label{tab:zeroatchars}
	\vspace{0.2cm}
	\centering
	\renewcommand{\arraystretch}{0.91}
	\begin{tabular}{lccc}
		\hline 
		& 
		\begin{tabular}[c]{@{}c@{}}\textbf{$\ell_2$-norm}\vspace{-0.15cm} \\ \textbf{reg.}\end{tabular} &
		\begin{tabular}[c]{@{}c@{}}\textbf{$\ell_1$-norm}\vspace{-0.15cm} \\ \textbf{reg.}\end{tabular} &
		\begin{tabular}[c]{@{}c@{}}\textbf{$\ell_0$-norm reg.}\vspace{-0.15cm} \\ \textbf{(proposed)}\end{tabular} \\ 
		\hline
		\begin{tabular}[c]{@{}l@{}}\textbf{Zero-attraction}\vspace{-0.15cm} \\ \textbf{term}\end{tabular}
		& $-2\eta\alpha w_j$ & $-\eta\alpha \, \mathrm{sign}(w_j)$ 
		& $-\eta\alpha \beta \mathrm{sign}(w_j) e^{-\beta|w_j|}$ \\ 
		\hline
		\begin{tabular}[c]{@{}l@{}}\textbf{Scaling factor}\vspace{-0.15cm} \\ \textbf{($\lambda_j = {w_j^{+}}/{w_j}$)}\end{tabular} 
		& $1 - 2 \eta \alpha $
		& $1 - \dfrac{\eta\alpha}{|w_j|}$
		& $1 - \dfrac{\eta\alpha\beta e^{-\beta |w_j|}}{|w_j|}$
		\\
		\hline
		\begin{tabular}[c]{@{}l@{}}\textbf{Bound of $\alpha$ for}\vspace{-0.15cm} \\ \textbf{weight shrinkage}\end{tabular} 
		& $ \alpha < \dfrac{1}{\eta} $
		& $ \alpha < \dfrac{2 |w_j|}{\eta} $
		& $ \alpha < \dfrac{2 |w_j|}{\eta \beta e^{-\beta |w_j|}}  $
		\\
		\hline
	\end{tabular} 	
\end{table}

To gain further insight on the differences between the norm-based regularization strategies, one interesting aspect to be analyzed is the scaling experienced by a weight after its update due to the zero attraction. To this end, we define a weight-update zero-attraction scaling factor $\lambda_j$, which is given by the ratio between $w_j^{+}$ and $w_j$ when only the zero-attraction term is considered for its update [i.e., ignoring $J(\mathbf{w}, \mathbf{x}, \mathbf{y})$ in either \eqref{eq:update_j_weight_2}, \eqref{eq:update_weights_l1}, or \eqref{eq:update_weight_aprox_l0_2}]. For instance, removing the last term on the right-hand side of \eqref{eq:update_weight_aprox_l0_2}, considering that $\mathrm{sign}(w_j) = w_j / |w_j|$, and manipulating the resulting expression, one can obtain the following zero-attraction scaling factor for the proposed $\ell_0$-norm-based regularization:
\begin{equation}\label{eq:lambdajl0}
	\lambda_j = \frac{w_j^{+}}{w_j} = 1 - \eta \alpha \beta \frac{e^{-\beta|w_j|}}{|w_j|}.
\end{equation}
By taking similar steps for the other norm-based regularization strategies, the scaling factors listed in Table~\ref{tab:zeroatchars} are obtained. 

The range of values of interest for the scaling factor $\lambda_j$ depends on the purpose of the regularization. In the case of overfitting mitigation, $\lambda_j$ should be smaller than and close to one, since this means that the weight is only mildly affected at each iteration. On the other hand, if the aim is weight pruning, the scaling factor should ideally be close to zero (stronger reduction of weight magnitude), but only for the weights to be pruned. In general terms, if \mbox{$-1 < \lambda_j < 1$}, the weight magnitude is reduced. Such an acceptable range for $\lambda_j$ can be used to  establish bounds for the choice of the penalization factor $\alpha$. For instance, using \eqref{eq:lambdajl0} in $-1 < \lambda_j < 1$ and manipulating the resulting expression, one obtains the following $\alpha$ bounds for the proposed $\ell_0$-norm-based regularization:
\begin{equation}\label{eq:alphaboundl0}
	0 < \alpha < \dfrac{ 2 |w_j| }{ \eta \beta e^{-\beta |w_j|} }.
\end{equation}
Since $\alpha$ is always positive, the lower bound in \eqref{eq:alphaboundl0} can be ignored. The upper bounds for $\alpha$ for the other norm-based regularization strategies are given in Table~\ref{tab:zeroatchars}, which are obtained in a similar way as \eqref{eq:alphaboundl0}. By analyzing such bounds, one can notice that, for both the $\ell_1$-norm and $\ell_0$-norm-based regularizations, the upper bound for $\alpha$ that ensures weight magnitude reduction is directly dependent on the weight magnitude $|w_j|$. Consequently, as the weight is attracted and approaches zero, such an upper bound will inevitably decrease and eventually be violated. At this point, we will have $|\lambda_j| > 1$ and a resulting growth of weight magnitude, corresponding to a zero repulsion instead of attraction. The weight growth will eventually increase the upper bound for $\alpha$, reverting the zero repulsion into attraction, starting the process towards repulsion again. Such an attraction-repulsion alternation will result in weight oscillations close to or around zero, which will be added to the noisy oscillations caused by the training algorithm itself after weight convergence (gradient noise). These oscillations may eventually prevent the weight magnitude from being mantained below the pruning threshold, resulting in the pruning of fewer weights. To mitigate this problem, $\alpha$ must be set to relatively small values taking into account also the adopted pruning threshold.

Now, in light of both the above discussion and the expressions given in Table~\ref{tab:zeroatchars}, some important conclusions can be drawn regarding the different norm-based regularization strategies as well as their zero-attraction capabilities. Considering first the $\ell_2$-norm regularization, from Table~\ref{tab:zeroatchars} we can notice that its $\alpha$ bound for weight magnitude reduction is irrespective of the weight magnitude. Consequently, the weight will always be attracted to zero without presenting the aforementioned zero-attraction-caused oscillations. However, since the strength of such an attraction becomes weaker as the weight becomes small, a smooth exponential weight evolution towards zero will be observed, as depicted in Figure~\ref{fig:zeroattractillustration}. As a consequence, too many training epochs are required for the weight to fall below the weight-pruning threshold. In summary, the zero-attraction provided by the $\ell_2$-norm regularization will be very interesting for avoiding overfitting, but not for weight pruning.

The $\ell_1$-norm regularization, in its turn, does not present the vanishing zero-attraction problem of the $\ell_2$-norm regularization, since its zero-attraction term is not dependent on the weight magnitude (see Table~\ref{tab:zeroatchars}). However, the $\ell_1$-norm regularization presents the aforementioned oscillations caused by the violation of the $\alpha$ bound, which will require $\alpha$ to be kept with relatively small values. This fact will inevitably harm the capability of the $\ell_1$-norm regularization to avoid the overfitting, since the zero attraction applied to larger weights is significantly limited by a small $\alpha$. In other words, the $\ell_1$-norm penalization has a problem of conflicting objectives, since $\alpha$ can be set either to overfitting avoidance or to favor zero attraction aiming at weight pruning.

\begin{figure}[!tb]
	\centering
	\includegraphics[scale=0.9]{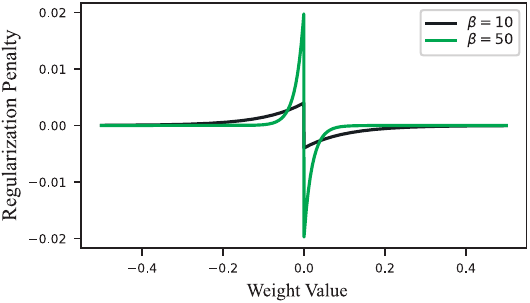}
	\vspace{-0.5cm}
	\caption{Penalization provided by the proposed $\ell_0$-norm regularization for two values of $\beta$.}
	\label{fig:exp_beta}
	\vspace{-0.3cm}
\end{figure}

The first aspect to highlight regarding the proposed $\ell_0$-norm based regularization is related to the direct dependence of its zero-attraction term on $e^{-\beta|w_j|}$ (see Table~\ref{tab:zeroatchars}). Since as $|w_j| \to \infty$, $e^{-\beta|w_j|} \to 0$, one can notice that the penalization tends to become very small for larger weights. Thus, the proposed $\ell_0$-norm regularization leans to be less effective in avoiding overfitting. On the other hand, in terms of zero-attraction for the smaller coefficients, the proposed $\ell_0$-norm regularization has a stronger effect that increases as the weight becomes smaller. This characteristic is illustrated in Figure~\ref{fig:exp_beta} for $\eta=0.04$ and $\alpha=0.01$, as well as for different values of $\beta$. As a result, during network training, a weight that becomes small is increasingly speeded up towards zero, taking a smaller number of epochs to reach the weight-pruning threshold. This behavior can be clearly observed in the curve corresponding to the $\ell_0$-norm penalization shown in Figure~\ref{fig:zeroattractillustration}. In terms of oscillations due to the zero-attraction, the $\ell_0$-norm-based regularization will also present such oscillations, which are due to the direct dependence on $|w_j|$ of its $\alpha$ bound for weight magnitude reduction. This aspect will require $\alpha$ to be kept with a relatively small value, with no major impact on the larger weights, since they are not affected by the $\ell_0$-norm-based regularization.

\subsection{Network Compression Scheme Based on Combined $\ell_2$-$\ell_0$-Norm Regularization}
From the discussion presented in the previous section, it becomes clear that the proposed $\ell_0$-norm-based regularization is very appropriate for obtaining networks having larger numbers of prunable weights, while presenting a poorer overfitting-avoidance capability. In contrast, the $\ell_2$-norm regularization is effective at avoiding overfitting, but not so for producing prunable weights. Thus, the idea here is to exploit such complementarity between the $\ell_2$- and $\ell_0$-norm-based regularizations to obtain highly-compressible networks that are also very effective in terms of inference performance. In this context, considering $\Omega(\mathbf{w}) = ||\mathbf{w}||_{2}^2 + ||\mathbf{w}||_{0}^*$ and taking the steps described in Section~\ref{sec:proposedl0} to derive the proposed $\ell_0$-norm regularization strategy, the following weight-update rule (training algorithm) is obtained:
\begin{equation}\label{eq:update_weight_combined}
w_j \leftarrow w_j - 2\eta\alpha_{\ell_2} w_j - \eta \alpha_{\ell_0} \beta \mathrm{sign}(w_j) e^{-\beta|w_j|} - \eta \frac{\partial J(\mathbf{w}; \mathbf{X}, \mathbf{y})}{\partial w_j}.
\end{equation}

Note that two different norm-penalization parameters ($\alpha_{\ell_2}$ and $\alpha_{\ell_0}$) are used in \eqref{eq:update_weight_combined} to  allow controlling the penalization level for each norm independently. This aspect is of great interest since, as previously mentioned, each norm penalization has a different task in the proposed training algorithm. Thus, adjusting $\alpha_{\ell_2}$, one can control the strength of the overfitting avoidance, whereas adjusting $\alpha_{\ell_0}$, the strength of zero attraction is controlled. Specifically regarding $\alpha_{\ell_0}$, such a parameter in fact allows controlling the trade-off between accuracy loss and compression level, which is typical in network compression techniques. In this context, using larger values of $\alpha_{\ell_0}$, compression is favored over maintaining the accuracy level of the uncompressed network, whereas smaller values of $\alpha_{\ell_0}$ lead to smaller compression levels while maintaining the original accuracy.

Finally, with \eqref{eq:update_weight_combined} at hand, we can define the proposed scheme for compression of neural networks, which is based on network training using the combined $\ell_2$-$\ell_0$-norm regularization followed by pruning the less significant weights of the resulting network. The proposed scheme is devised with unstructured (weight) pruning in mind, since the penalization produces a zero-attraction for each weight independently. Despite this, structured (neuron) pruning can also be attained via a clever design of the pruning strategy. In the next section, the effectiveness of the proposed scheme is assessed.

	%
	
	\vspace{-0.2cm}
\section{Experimental Results}\label{sec:results}

This section is focused first on assessing the ability of the proposed $\ell_0$-norm regularization approach on inducing sparseness during network training and then on evaluating the performance of the proposed network compression scheme based on combined $\ell_2$-$\ell_0$-norm regularization. In this context, six different experiments are considered. In the first, the idea is to compare the sparseness induction capability of the proposed $\ell_0$-norm regularization with those of the more standard $\ell_2$- and $\ell_1$-norm regularizations, as well as to evaluate their impact when distinct pruning strategies are used. In the second experiment, the proposed network compression scheme is used for obtaining compressed versions of the Lenet-300-100 and Lenet-5-Caffe networks applied to the MNIST dataset. 
The third experiment involves the application of the proposed network compression scheme to the task of compressing the VGG-16 network and also a residual network called ResNet20, both applied to the CIFAR-10 dataset. In the forth experiment, the focus is on compressing the VGG-16 network applied to the more challenging task of the CIFAR-100, whereas the fifth experiment uses the same dataset, but now considering the ResNet50 network with a limited number of training epochs. Finally, in the sixth and last experiment, structural pruning is used together with the proposed regularization technique.

In general, the networks are assessed in terms of their accuracy versus compression ratio trade-off. Aiming to facilitate such an assessment, curves of accuracy versus compression ratio are presented for the proposed approach in most of the experiments. Comparisons with other results from the open literature are also presented, considering the accuracy and compression values reported in the respective competing papers. Since the accuracy versus compression trade-offs varies slightly from one paper to the other, we aim at a fair comparison by considering similar accuracy levels when choosing the trade-off for the proposed approach. Moreover, in some cases, different results are presented for the proposed approach, corresponding to different trade-offs.

\subsection{Experiment 1: Sparseness Induction Comparison Between Different Norm-Based Regularization Strategies}

The aim of this experiment is to assess the sparseness-induction capability of the proposed $\ell_0$-norm regularization as compared with both the standard $\ell_2$- and $\ell_1$-norm-based regularizations. In this context, neural networks inspired on the Lenet-5-Caffe\footnote{The Lenet-5-Caffe is a modified version of Lenet-5 \cite{lecun1998gradient} which is discussed in \cite{jia2014caffe}.} are trained for 
the image classification task of the MNIST dataset. To assess the trade-off between accuracy and network compression ratio, the networks obtained after training are increasingly compressed (starting with a factor of 2) and the resulting accuracy is evaluated for each compression ratio. The network compression is carried out by using three different weight-pruning strategies \cite{blalock2020state}: i) global magnitude pruning (GP), where the less significant weights of the network as a whole are targeted for pruning; ii) layerwise magnitude pruning (LP), where the less significant weights in each layer are targeted (resulting in the same compression ratio for each layer); and iii) random pruning (RP), where weights are randomly selected from the whole network, regardless of their magnitude. Fine tuning (FT) \cite{lecun1998gradient} (i.e., network retraining after pruning) is also considered in this experiment. 

\begin{figure}[!b]
	\centering
	\includegraphics[scale=0.95]{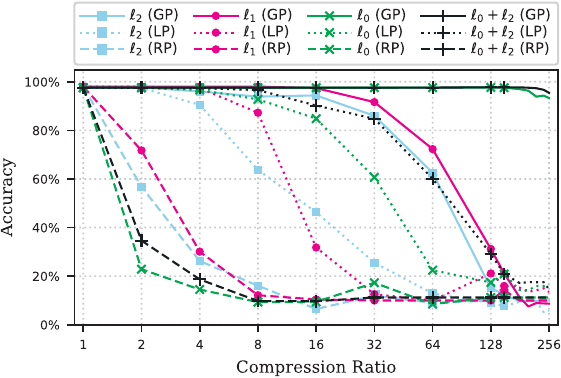}
	\vspace{-0.5cm}
	\caption{Experiment 1. Accuracy versus compression ratio without FT.}
	\label{fig:compare_regularizations_wo_fine_tuning}
	\vspace{-0.2cm}
\end{figure}

\begin{figure}[!t]
	\centering
	\includegraphics[scale=0.95]{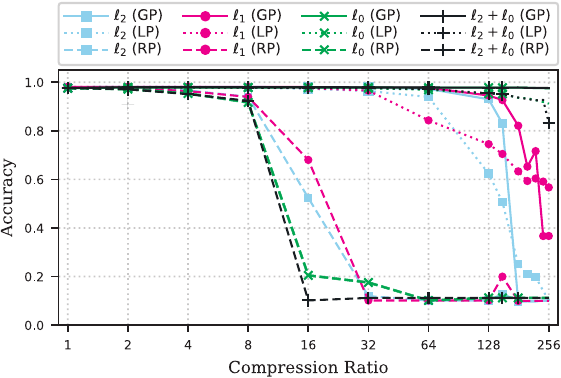}
	\vspace{-0.5cm}
	\caption{Experiment 1. Accuracy versus compression ratio with FT.}
	\label{fig:compare_regularizations}
	\vspace{-0.2cm}
\end{figure}

Figures~\ref{fig:compare_regularizations_wo_fine_tuning} and \ref{fig:compare_regularizations} present the curves of accuracy versus compression ratio obtained without and with FT, respectively. The first aspect to highlight, from these figures, is the superiority of both global and layerwise weight pruning with respect to random pruning. This fact demonstrates the effectiveness of network compression via magnitude-based weight pruning. Now, regarding the capability of the different norm-based regularization techniques to induce sparseness and thus provide higher compression ratios, the curves from Figures \ref{fig:compare_regularizations_wo_fine_tuning} and \ref{fig:compare_regularizations} show that the proposed $\ell_0$-norm-based regularization (both alone and in combination with the $\ell_2$ norm) significantly outperforms the regularizations based on both $\ell_1$ and $\ell_2$ norms. This conclusion is direct from the superior results obtained by using the proposed regularization with either global or layerwise weight pruning. Moreover, the inferior performance obtained by the proposed regularization with random pruning is also due to its superior sparseness inducing capability, since a higher sparseness level  implies more concentration of energy in fewer weights, with some of these more significant weights inevitably targeted by a random pruning strategy. Another important aspect to point out is that the use of FT leads to significantly better results in all cases, even though the proposed $\ell_0$-norm-based regularization has obtained very good results even without using FT (see Figure~\ref{fig:compare_regularizations_wo_fine_tuning}).

\subsection{Experiment 2: Network Compression for the MNIST Dataset}

In this experiment, we consider the MNIST dataset and compare other  pruning-based network compression approaches from the literature with the network compression scheme proposed in the present paper. In this context, two network topologies are considered, namely, Lenet-300-100 and Lenet-5-Caffe. The chosen pruning strategy is the global pruning due to the superior results observed in the first experiment.
For setting the training hyperparameters, we consider two different strategies. The first, termed NORM, is based on using layer-size normalized values for $\alpha_{\ell_2}$ and $\alpha_{\ell_0}$ \cite{li2019l0}. Thus, three global hyperparameter values ($\alpha_{\ell_2}$, $\alpha_{\ell_0}$, and $\beta$) are tuned experimentally, besides the learning rate $\eta$. In the second strategy, called SEP, individual values of $\alpha_{\ell_2}$, $\alpha_{\ell_0}$, and $\beta$ are used for each network layer. In this case, we start with the hyperparameter values from the NORM strategy and fine tune individual hyperparameters for each layer experimentally. This procedure allows enhancing the penalization in layers possessing a higher number of weights with larger magnitudes without compromising the overall network accuracy. In general terms, NORM is simpler to use, whereas SEP tends to produce better results due to the increased degrees of freedom.

\begin{figure}[!t]
	\vspace{-0.9cm}
	\centering
	\includegraphics[scale=0.95]{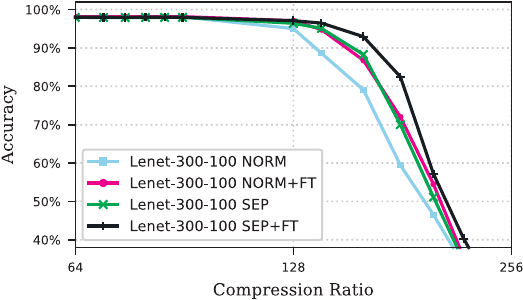}
	\vspace{-0.6cm}
	\caption{Experiment 2. Accuracy versus compression ratio obtained applying the proposed scheme to the LeNet-300-100 network.}
	\label{fig:Lenet-300-100-CompressionRate}
\end{figure}

\begin{figure}[!t]
	\vspace{-0.2cm}
	\centering
	\includegraphics[scale=0.95]{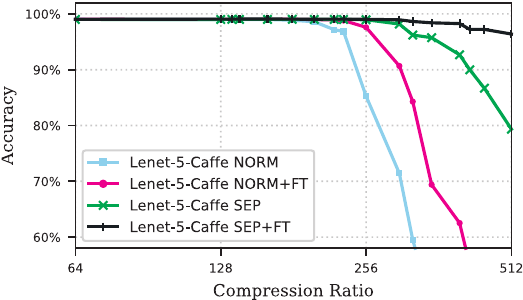}
	\vspace{-0.6cm}
	\caption{Experiment 2. Accuracy versus compression ratio obtained applying the proposed scheme to the LeNet-5-Caffe network.}
	\label{fig:Lenet5-CompressionRate}
	\vspace{-0.5cm}
\end{figure}

The first aspect evaluated in this experiment is the trade-off between accuracy and compression ratio obtained by using the proposed approach. Figures \ref{fig:Lenet-300-100-CompressionRate} and \ref{fig:Lenet5-CompressionRate} depict such a trade-off for the Lenet-300-100 and Lenet-5-Coeffee networks, respectively, considering both NORM and SEP strategies, as well as FT. From these figures, one can notice that very high compression ratios can be obtained by using the proposed approach with negligible performance loss for the cases considered here. As expected, the best results are obtained by using the combination of SEP strategy with FT (around $100\times$ compression ratio for Lenet-300-100 and more than $300\times$ for Lenet-5-Caffe). However, even the simpler NORM strategy was capable of leading to satisfactory values (around $90\times$ for Lenet-300-100 and $200\times$ for Lenet-5-Caffe).  

The results obtained by using the proposed approach are now confronted with those obtained by using the following pruning-based compression methods from the literature: threshold pruning (THP) \cite{NIPS2015_5784}, dynamic network surgery (DNS) \cite{DBLP:journals/corr/GuoYC16}, group horseshoe (BC-GHS) \cite{louizos2017bayesian}, sparse variational dropout (SparseVD) \cite{molchanov2017variational}, cutting plane algorithm (CPA) \cite{phan2020pruning}, global sparse momentum (GSM) \cite{ding2019global}, and automatic network pruning (Autoprune) \cite{xiao2019autoprune}. Moreover, we also consider the following pruning methods based on the use of the $\ell_0$ norm:
 $\ell_0$ with hard-concrete approximation ($L_0$-HC) \cite{louizos2017learning}, $\ell_0$ augment-reinforce-merge ($L_0$-ARM) \cite{li2019l0}, $\ell_0$ based on ADMM ($L_0$-ADMM) \cite{zhangECCV2018}, and learn compression using $\ell_2$ and $\ell_0$ norms ($L_0$-$L_2$-LC) \cite{cperpinanL0L2} (results from \cite{cperpinanL0} are not included since the approach in  \cite{cperpinanL0L2} is a direct evolution of that in \cite{cperpinanL0} and presents superior results).
The comparisons among all approaches considered are presented in Tables \ref{tab:resultados_poda_iterativa} and \ref{tab:resultados_poda_iterativa2} for LeNet-300-100 and LeNet-5-Caffe, respectively. From these tables, one can infer that the proposed approach is very competitive, generally resulting in superior trade-offs between accuracy and compression ratio as compared with the other approaches from the literature.

\begin{table}[!t]
	\vspace{-0.5cm}
	\caption{Summary of results for the LeNet-300-100 architecture on the MNIST dataset}
	\label{tab:resultados_poda_iterativa}
	\centering
	\renewcommand{\arraystretch}{0.825}
	\vspace{0.2cm}
	\begin{tabular}{lccc} 				
		\hline 
		\begin{tabular}[c]{@{}l@{}}\textbf{Prune}\vspace{-0.15cm} \\ \textbf{Method} \end{tabular} & 
		\begin{tabular}[c]{@{}c@{}}\textbf{Top-1}\vspace{-0.15cm} \\ \textbf{Error}\end{tabular} &
		\begin{tabular}[c]{@{}c@{}}\textbf{Total}\vspace{-0.15cm}\\\textbf{Parameters}\end{tabular} & 
		\begin{tabular}[c]{@{}c@{}}\textbf{Compression}\vspace{-0.15cm}\\ \textbf{Ratio}\end{tabular} \\ \hline
		No Prune                                                & $1.64\%$                                               & 267k                                                   & -                                                          \\
		THP \cite{NIPS2015_5784}                                    & $1.59\%$                                               & 22k                                                    & $12 \times$                                                \\
		DNS \cite{DBLP:journals/corr/GuoYC16}                       & $1.99\%$                                               & 4.8k                                                   & $56 \times$                                                \\
		BC-GHS \cite{louizos2017bayesian}                              & $1.8\%$                                               & 29.6k                                                    & $9 \times$                                                \\
		SparseVD \cite{molchanov2017variational}                         & $1.92\%$                                               & 3.92k                                                    & $68 \times$                                                \\
		CPA  \cite{phan2020pruning}                        & $1.77\%$                                               & 3.14k                                                    & $85 \times$                                                \\
		GSM \cite{ding2019global}                         & $1.82\%$                                               & 4.45k                                                    & $60 \times$                                                \\		
		Autoprune \cite{xiao2019autoprune}                         & $1.78\%$                                               & 3.34k                                                    & $80 \times$                                                \\
		\hline
		$L_0$-HC \cite{louizos2017learning}                             & $1.8\%$                                               & 26.72k                                                    & $9.99 \times$                                                \\
		$L_0$-ARM \cite{li2019l0}                                         & $1.9\%$                                               & 18.8k                                                    & $14.2 \times$                                                \\
		$L_0$-ADMM \cite{zhangECCV2018}                                         &   $^*$                                              & 11.6k                                                    & $23 \times$                                                \\
		$L_0$-$L_2$-LC  \cite{cperpinanL0L2}                                        & $1.6\%$                                              & 5.34k                                                    & $50 \times$                                                \\
		$L_0$-$L_2$-LC  \cite{cperpinanL0L2}                                        & $1.99\%$                                               & 4k                                                    & $66.8 \times$                                                \\	
		\hline			
        \textbf{Proposed (NORM)}            & $1.85\%$                                               & 2.96k                                                   & $90 \times$                                                \\
        \textbf{Proposed (SEP)}            & $1.99\%$                                               & 2.7k                                                   & $96 \times$                                                \\ \hline
        \multicolumn{4}{l}{\footnotesize $^*$ The same as No Prune.}
	\end{tabular} 
	
\end{table}

\begin{table}[!b]
	\vspace{-0.3cm}
	\caption{Summary of results for the LeNet-5-Caffe architecture on the MNIST dataset}
	\label{tab:resultados_poda_iterativa2}
	\centering
	\renewcommand{\arraystretch}{0.825}
	\vspace{0.2cm}	
\begin{tabular}{lccc} 	
 \hline 
 \begin{tabular}[c]{@{}l@{}}\textbf{Prune}\vspace{-0.15cm} \\ \textbf{Method} \end{tabular} & 
 \begin{tabular}[c]{@{}c@{}}\textbf{Top-1}\vspace{-0.15cm} \\ \textbf{Error}\end{tabular} &
 \begin{tabular}[c]{@{}c@{}}\textbf{Total}\vspace{-0.15cm}\\\textbf{Parameters}\end{tabular} & 
 \begin{tabular}[c]{@{}c@{}}\textbf{Compression}\vspace{-0.15cm}\\ \textbf{Ratio}\end{tabular} \\ \hline 
    No Prune                                                & $0.8\%$                                                & 431k                                                   &   -                                                         \\
	THP \cite{NIPS2015_5784}                                    & $0.7\%$                                                & 36k                                                    & $12 \times$                                                \\
	DNS \cite{DBLP:journals/corr/GuoYC16}                       & $0.91\%$                                               & 4.0k                                                   & $108\times$                                                \\
	BC-GHS \cite{louizos2017bayesian}                               & $1.0\%$                                               & 2.76k                                                    & $156 \times$                                                \\			
	SparseVD  \cite{molchanov2017variational}                        & $0.75\%$                                               & 1.54k                                                    & $280 \times$                                                \\
	CPA  \cite{phan2020pruning}                        & $0.79\%$                                               & 1.35k                                                    & $317 \times$                                                \\
	GSM \cite{ding2019global}                         & $0.94\%$                                               & 1.44k                                                    & $300 \times$                                                \\		
	Autoprune \cite{xiao2019autoprune}                         & $0.91\%$                                               & 1.39k                                                    & $310 \times$                                                \\
	\hline
	$L_0$-HC \cite{louizos2017learning}                               & $1.0\%$                                               & 4.66k                                                    & $92.6 \times$                                                \\
	$L_0$-ARM  \cite{li2019l0}                                     & $1.3\%$                                               & 2.2k                                                    & $196 \times$                                                \\
	$L_0$-ADMM  \cite{zhangECCV2018}                                       &  $^*$                                              & 6.05k                                                    & $71.2 \times$                                                \\
	$L_0$-$L_2$-LC \cite{cperpinanL0L2}                                         & $0.89\%$                                               &  4.3k                                                   & $ 100 \times$                                                \\				
	\hline
	\textbf{Proposed (NORM)}            & $0.99\%$                                               & 2.15k                                                  & $200\times$                                                \\
	\textbf{Proposed (SEP)}             & $0.97\%$                                               & 1.2k                                                  & $359\times$                                                \\ \hline
	\multicolumn{4}{l}{\footnotesize $^*$ The same as No Prune.}
	\end{tabular}
\end{table}

\clearpage

\subsection{Experiment 3: CIFAR-10}
For this third experiment, we consider the image classification task of the CIFAR-10 dataset using both the VGG-16 \cite{simonyan2014very} and ResNet20 \cite{He2016} networks. The VGG-16 has around 15 million parameters and typically obtains an error rate of $6.75\%$ for the CIFAR-10. The ResNet20 belongs to the family of residual networks (ResNets), in which residual connections are exploited to obtain smaller computational footprints. Thus, this network is implemented with 22 layers and about 274 thousand parameters, obtaining an error rate of $7.8\%$ for the CIFAR-10 dataset. In this experiment, these two architectures are considered aiming to assess the capability of the proposed compression scheme for dealing with larger networks (such as the VGG-16) as well as with a residual one (ResNet20) for which network compression tends to be a more challenging task due to the already reduced number of parameters.

\begin{table}[!b]	
	\vspace{-0.25cm}
	\caption{Summary of results for the VGG-16 architecture applied to the CIFAR-10 dataset}
	\label{tab:vgg16cifar10}
	\centering
	\renewcommand{\arraystretch}{0.825}
	\begin{tabular}{cccc}
		\hline
		\multicolumn{1}{c}{\textbf{Prune Method}} & \multicolumn{1}{c}{\begin{tabular}[c]{@{}c@{}}\textbf{Error} \\ \textbf{Increased}\end{tabular}} & \multicolumn{1}{c}{\begin{tabular}[c]{@{}c@{}}\textbf{Parameters}\\ \textbf{Dropped}\end{tabular}} & \multicolumn{1}{c}{\begin{tabular}[c]{@{}c@{}}\textbf{Compression} \\ \textbf{Ratio}\end{tabular}} \\ \hline
		SSS \cite{huangwang2018}                          & $^*$                                                                          & $30\%$                                                                        & $1.67\times$                                                                            \\	     
		EConvNets \cite{LiKadav2017}                          & $-0.15\%$                                                                         & $64\%$                                                                        & $2.78\times$                                                                            \\
		VAR-CONV \cite{Zhao2019}                       & $0.07\%$                                                                         & $73.34\%$                                                                        & $3.75\times$                                                                            \\
		PP \cite{singh2019}                           & $0.03\%$                                                                          & $82.8\%$                                                                        & $5.8\times$                                                                            \\	
		PP \cite{singh2019}                           & $0.14\%$                                                                          & $84.4\%$                                                                        & $6.43\times$  		                                                                           \\		
		RFP \cite{AYINDE2019148}                         & $0.13\%$                                                                         & $78.1\%$                                                                        & $4.6\times$                                                                            \\
		RFP \cite{AYINDE2019148}                            & $0.72\%$                                                                          & $89.5\%$                                                                        & $9.52\times$                                                                            \\ \hline
		\textbf{Proposed (NORM)}                            & $0.41\%$          & $85.5\%$                                                                           & $6.92\times$                                                                               \\
		\textbf{Proposed (NORM)}                            & $0.7\%$          & $93.2\%$                                                                           & $14.76\times$                                                                               \\
		\hline
		\multicolumn{4}{l}{\footnotesize $^*$ The same as No Prune.}
	\end{tabular}
\vspace{-0.25cm}
\end{table}

For the case of the VGG-16 network, the NORM strategy is used for adjusting the hyperparameters and FT is also used. The SEP strategy was not used in this case to avoid the adjustment of too many hyperparameters. The obtained results are shown in Table \ref{tab:vgg16cifar10} along with results from other competing approaches from the literature. From this table, one can notice that the proposed scheme is capable of obtaining very good results even when the SEP is not used, outperforming the competing approaches with an acceptable level of error increase.  

\begin{figure}[!bt]
	\centering
	\includegraphics[scale=0.95]{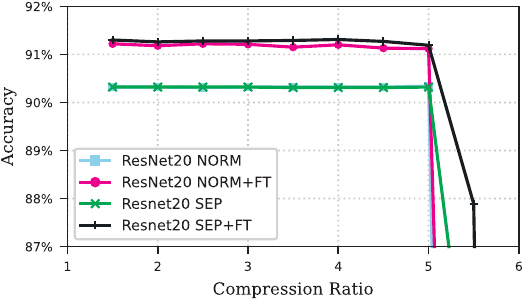}
	\vspace{-0.5cm}
	\caption{Experiment 3. Accuracy versus compression ratio obtained applying the proposed scheme to the ResNet20 network.}
	\label{fig:ResNet20-Cifar10}
\end{figure}

For the case of the ResNet20, due to the smaller complexity of such a network, we consider both NORM and SEP strategies for adjusting the hyperparameters of the proposed compression scheme, as well as FT. The curves of accuracy versus compression ratio obtained by using the different combinations of these strategies are presented in Figure~\ref{fig:ResNet20-Cifar10}. These results demonstrate the effectiveness of the proposed scheme applied to ResNets, leading to relatively high compression ratios (up to 5 times) with acceptable performance loss. One important aspect to highlight in this case is that the difference between NORM and SEP strategies is very small (contrasting with the more significant difference observed in Experiment 2), whereas FT has resulted in more important performance gains. The comparison with other works from the literature is presented in Table~\ref{tab:cifar10}. From this table, one observes that the proposed scheme outperforms the other approaches from the open literature, demonstrating again the effectiveness of the association of combined $\ell_2$-$\ell_0$-norm regularization and weight pruning.

\clearpage

\begin{table}[!htb]	
	\caption{Summary of results for the ResNet-20 architecture applied to the CIFAR-10 dataset}
	\label{tab:cifar10}
	\centering
	\renewcommand{\arraystretch}{0.825}
	\begin{tabular}{cccc}
		\hline
		 \multicolumn{1}{c}{\textbf{Prune Method}} & \multicolumn{1}{c}{\begin{tabular}[c]{@{}c@{}}\textbf{Error} \\ \textbf{Increased}\end{tabular}} & \multicolumn{1}{c}{\begin{tabular}[c]{@{}c@{}}\textbf{Parameters}\\ \textbf{Dropped}\end{tabular}} & \multicolumn{1}{c}{\begin{tabular}[c]{@{}c@{}}\textbf{Compression} \\ \textbf{Ratio}\end{tabular}} \\ \hline
		VAR-CONV \cite{Zhao2019}                       & $0.54\%$                                                                         & $20.41\%$                                                                        & $1.26\times$                                                                            \\
		CNN-FCF \cite{Li2019}                          & $1.07\%$                                                                         & $42.75\%$                                                                        & $1.75\times$                                                                            \\
		CNN-FCF \cite{Li2019}                         & $2.67\%$                                                                         & $68.44\%$                                                                        & $3.17\times$                                                                            \\
		SNLI \cite{Ye2018}                            & $1.1\%$                                                                          & $37.22\%$                                                                        & $1.59\times$                                                                            \\
		SNLI \cite{Ye2018}                           & $3.2\%$                                                                          & $67.83\%$                                                                        & $3.11\times$                                                                            \\
		DHP \cite{li2020dhp}                           & $1.0\%$                                                                          & $55.95\%$                                                                        & $2.27\times$                                                                            \\  \hline		
		Proposed (NORM)                            & $1.1\%$                                                                        & $80\%$                                                                           & $5\times$                                                                               \\
		Proposed (SEP)                           & $1\%$                                                                          & $80\%$                                                                        & $5\times$                                                                               \\ \hline
	\end{tabular}
\end{table}

\subsection{Experiment 4: VGG-16 on CIFAR-100}

In this forth experiment, the VGG-16 network is applied to the classification task of the CIFAR-100 dataset, which comprises 100 different classes in contrast to the 10 different classes of the CIFAR-10. Thus, one can notice that the VGG-16 is now subject to a more challenging task, resulting in accuracy levels on the order of $70\%$ (against more than $90\%$ for the CIFAR-10). Such an increased difficulty also results in a more challenging network compression task. For the sake of simplicity, FT is not used in this case and only the NORM strategy is used for adjusting the parameters of the proposed scheme. Moreover, four different values of $\alpha_{\ell_0}$ are considered: 0 (i.e., without $\ell_0$-norm penalization), $1 \times 10^{-7}$, $2.5 \times 10^{-7}$, and $5 \times 10^{-7}$. The idea is to demonstrate how the trade-off between accuracy and compression ratio can be controlled via adjusting $\alpha_{\ell_0}$.  

\begin{figure}[!bt]
	\centering
	\includegraphics[scale=0.95]{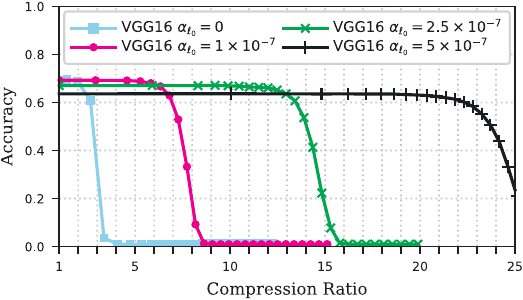}
	\vspace{-0.3cm}
	\caption{Experiment 4. Accuracy versus compression ratio obtained applying the proposed scheme to the VGG-16 network with the CIFAR-100 dataset.}
	\label{fig:VGG16-Cifar100}
\end{figure}

Figure~\ref{fig:VGG16-Cifar100} shows the curves of accuracy versus compression ratio obtained by using the different values of $\alpha_{\ell_0}$. From these curves, one can notice that, as the value of $\alpha_{\ell_0}$ grows (i.e., the regularization effect is strengthened), the compression ratio that can be attained without significant accuracy loss is increased. This comes at the cost of slightly reducing the initial accuracy level (i.e., the one obtained before pruning). More specifically, the initial accuracy obtained for $\alpha_{\ell_0} = 0$ is reduced by $0.19\%$, $2.42\%$, and $5.85\%$ for the values of $\alpha_{\ell_0}$ of $1 \times 10^{-7}$, $2.5 \times 10^{-7}$, and $5 \times 10^{-7}$, respectively. 

Table~\ref{table:cifar100} describes the compression results obtained by the proposed scheme for each of the considered values of $\alpha_{\ell_0}$ along with the results from \cite{Zhao2019}. Another result from a somewhat similar setup from the literature can be found in \cite{wangsensors2020}. The method proposed there is capable of reducing the number of parameters by $81.1\%$ for the VGG-16 in CIFAR-100, corresponding to a compression ratio of $5.29\times$, at the cost of an increase in the error rate of $0.17\%$. This result is comparable with those obtained by using the scheme proposed here. However, the VGG-16 version considered there has more than twice the size of the VGG-16 considered here and thus, after compression, it still has 6.43 million parameters, whereas in our case the compressed network with $\alpha_{\ell_0} = 1 \times 10^{-7}$ has around 3.2 million parameters. From all these results, one can notice again the effectiveness of the proposed scheme.

\begin{table}[!tb]
	\renewcommand{\tabcolsep}{4pt}
	\caption{Summary of results for the VGG-16 network applied to the CIFAR-100 dataset}
	\label{table:cifar100}
	\centering
	\renewcommand{\arraystretch}{0.825}
	\begin{tabular}{cccc}
		
		\hline
		\multicolumn{1}{c}{\textbf{Prune Method}} & \multicolumn{1}{c}{\begin{tabular}[c]{@{}c@{}}\textbf{Error} \\ \textbf{Increased}\end{tabular}} & \multicolumn{1}{c}{\begin{tabular}[c]{@{}c@{}}\textbf{Parameters}\\ \textbf{Dropped}\end{tabular}} & \multicolumn{1}{c}{\begin{tabular}[c]{@{}c@{}}\textbf{Compression} \\ \textbf{Ratio}\end{tabular}} \\ \hline
		VAR-CONV \cite{Zhao2019}                       & $-0.07\%$                                                                         & $37.87\%$                                                                        & $1.61\times$                                                                            \\	
		Proposed with $\alpha_{\ell_0} = 1 \times 10^{-7}$                           & $0.3\%$                                                                          & $78.6\%$                                                                        & $4.67\times$                                                                               \\ 
		Proposed with $\alpha_{\ell_0} = 2.5 \times 10^{-7}$                           & $2.67\%$                                                                          & $90.4\%$                                                                        & $10.47\times$                                                                               \\
		Proposed with $\alpha_{\ell_0} = 5 \times 10^{-7}$                           & $5.95\%$                                                                          & $94.6\%$                                                                        & $18.63\times$                                                                               \\\hline
	\end{tabular}
\end{table}

\clearpage

\subsection{Experiment 5: ResNet-50 on CIFAR-100 with Limited Number of Epochs}
The fifth experiment of this work is dedicated to evaluate the performance of the proposed approach for the compression of a ResNet-50 model \cite{He2016} on CIFAR-100 considering a limited number of 100 training epochs. The idea is to evaluate the capability of the proposed approach for producing a sparse compressible network without too much training. As a reference, the accuracy attained by the considered model after 100 training epochs without using the proposed regularization was around $71.5\%$. Again, for the sake of simplicity, FT is not used and the NORM strategy is applied for adjusting the hyperparameters of the proposed scheme. Moreover, two different values of $\beta$ (5 and 10) and three different values of $\alpha_{\ell_0}$ ($1 \times 10^{-7}$, $5 \times 10^{-7}$, and $1 \times 10^{-6}$) are considered.

\begin{figure}[!b]
	\centering
	\includegraphics{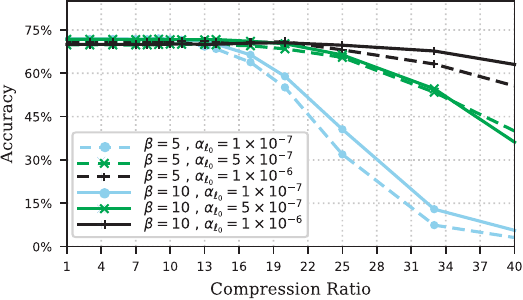}
	\vspace{-0.4cm}
	\caption{Experiment 5. Accuracy versus compression ratio obtained applying the proposed scheme to the ResNet-50 network with the CIFAR-100 dataset.}
	\label{fig:ResNet50-Cifar100}
\end{figure}

The obtained curves of accuracy versus compression ratio are presented in Figure~\ref{fig:ResNet50-Cifar100}. From these curves, one can again notice the effectiveness of the proposed approach, producing compression rates varying from $13 \times$ (with $\beta=5$ and $\alpha_{\ell_0}=1 \times 10^{-7}$) to $25 \times$ (with $\beta=10$ and $\alpha_{\ell_0}=1 \times 10^{-6}$) without noticeable performance reduction. Moreover, one can observe that the change in the value of $\beta$ from 5 to 10 does not result in a severe performance change, showing a certain robustness of the proposed strategy with respect to the choice of such a parameter.

\clearpage

\subsection{Experiment 6: Structural Pruning}
In this sixth experiment, the effectiveness of the proposed regularization approach applied to structural pruning is assessed. As in Experiment 3, the architecture considered is VGG-16 and the dataset is CIFAR-10. Two network pruning strategies are considered: i) neuron pruning (NP), in which we prune the neurons (dense-layer neurons or convolutional-layer filters) whose weight vectors present the smallest $\ell_2$ norms; and ii) NP combined with global weight pruning (NP+GP). These strategies are compared in terms of both compression ratio and reduction in the number of floating-point operations (FLOPs) required for inference. In the case of the FLOPs comparison, we consider the reduction exclusively due to neuron pruning (implying that, in the NP+GP case, we only consider the FLOPs reduction due to NP). Table~\ref{table:cifar10-struc-prun} presents the results obtained by using the proposed compression scheme along with the findings from \cite{LiKadav2017}.
Such results show that the proposed compression strategy holds the potential to be also used for neuron pruning, as the compression levels obtained are competitive with those attained by an approach centered on structural  pruning \cite{LiKadav2017}. Moreover, when NP is combined with GP, the reduction in FLOPs is again very good, whereas the attained network compression levels (which account also for the pruning of individual weights) are very high indeed.

\begin{table}[!b]
	\vspace{-0.5cm}
	\caption{Summary of results for structural pruning of the VGG-16 architecture applied to the CIFAR-10 dataset}
	\label{table:cifar10-struc-prun}
	\centering
	\vspace{0.1cm}
	\renewcommand{\arraystretch}{0.825}
	\begin{tabular}{cccc}		
		\hline
		\multicolumn{1}{c}{\textbf{Prune Method}} & \multicolumn{1}{c}{\begin{tabular}[c]{@{}c@{}}\textbf{Error} \\ \textbf{Increased}\end{tabular}} & \multicolumn{1}{c}{\begin{tabular}[c]{@{}c@{}}\textbf{FLOP}\\ \textbf{Reduction}$^*$\end{tabular}} & \multicolumn{1}{c}{\begin{tabular}[c]{@{}c@{}}\textbf{Compression} \\ \textbf{Ratio}\end{tabular}} \\ \hline
		EConvNets \cite{LiKadav2017}                       & $-0.15\%$                                                                         & $34.2\%$                                                                        & $2.78\times$                                                                            \\	
		Proposed (NP)                           & $0.31\%$                                                                          & $27.75\%$                                                                        & $1.46\times$                                                                               \\ 
		Proposed (NP)                           & $0.49\%$                                                                          & $42.7\%$                                                                        & $1.7\times$                                                                               \\
		Proposed (NP+GP)                           & $0.27\%$                                                                          & $27.75\%$                                                                        & $9.21\times$                                                                               \\		
		Proposed (NP+GP)                          & $0.72\%$                                                                          & $42.7\%$                                                                        & $18.38\times$                                                                               \\\hline
		\multicolumn{4}{l}{\footnotesize $^*$ Exclusively due to neuron pruning.}
	\end{tabular}
\end{table}

\color{black}
	%
	
	\section{Concluding Remarks}
\label{sec:conc}
The focus of this paper was on the development of a novel neural-network compression scheme based on sparseness-inducing training followed by weight pruning. In this context, a new $\ell_0$-norm-based regularization approach was introduced aiming to strengthen the sparseness induction during training and thus obtain effective networks with larger numbers of prunable weights. Such a regularization approach is however not as effective in avoiding model overfitting. Thus, a combined  $\ell_2$-$\ell_0$-norm regularization is considered for the implementation of the proposed network compression scheme. A set of experimental results is presented, confirming the effectiveness of the proposed scheme in scenarios involving fully-connected, convolutional, and residual networks.  

\section*{Acknowledgments}
The authors are thankful to the Handling Editor and the anonymous reviewers whose valuable comments and constructive suggestions have significantly benefited this paper.

This research work was supported in part by the Coordination for the Improvement of Higher Education Personal (CAPES) and the Brazilian National Council for Scientific and Technological Development (CNPq) from Brazil.

	\bibliography{sparse-nets-refs}
	
\end{document}